\documentclass[lettersize,journal]{IEEEtran}
\usepackage{amsmath,amsfonts}
\usepackage{algorithm}
\usepackage{array}
\usepackage[caption=false,font=normalsize,labelfont=sf,textfont=sf]{subfig}
\usepackage{textcomp}
\usepackage{stfloats}
\usepackage{url}
\usepackage{verbatim}
\usepackage{graphicx}
\usepackage{cite}
\usepackage{cite}
\usepackage{booktabs}
\usepackage{multirow}
\usepackage{graphicx} 
\usepackage{hyperref}
\usepackage{makecell}
\usepackage{mathtools}
\usepackage{algpseudocode} 
\usepackage{tabularx}

\usepackage{color}
\usepackage[table]{xcolor} 
\usepackage{array}

\newcolumntype{C}{>{\centering\arraybackslash}X}

\begin{document}

\title{\huge Bridging the Gap between Labeled and Unlabeled Data via Unified Flow with Feature Memory Bank}

\author{\IEEEauthorblockN{Shanwen Wang,~\IEEEmembership{Graduate Student Member,~IEEE}, Xin Sun,~\IEEEmembership{Senior Member,~IEEE}, Danfeng Hong,~\IEEEmembership{Senior Member,~IEEE}, Junyu Dong, \IEEEmembership{Member,~IEEE}, Patrick Le Callet,~\IEEEmembership{Fellow,~IEEE}
}
\thanks{This work is supported by the Science and Technology Development Fund - International Collaborative Research, Macao SAR (0001/2025/AIJ), Science and Technology Development Fund, Macao SAR - Ministry of Science and Technology: National Key R\&D Program of China (0007/2025/AMJ, 2025YFE0202900), and Science and Technology Development Fund, Macao SAR - Basic Research (0006/2024/RIA1)} 
\thanks{S. Wang and X. Sun are with Faculty of Data Science, City University of Macau, 999078, SAR Macao, China. D. Hong is with School of Automation, Southeast University, Nanjing, 211189, China. J. Dong are with the Department of Computer Science and Technology, Ocean University of China, Qingdao, 266100 China. P.L. Callet is with Nantes Université, Ecole Centrale Nantes, CAPACITES SAS, CNRS, LS2N, UMR 6004, Nantes, France.}

}

\markboth{Journal of \LaTeX\ Class Files,~Vol.~14, No.~8, August~2021}%
{Shell \MakeLowercase{\textit{et al.}}: A Sample Article Using IEEEtran.cls for IEEE Journals}


\maketitle

\begin{abstract}
Although semi-supervised semantic segmentation ($\text{S}^4$) utilizes abundant unlabeled data to reduce manual labeling burdens, independent training of labeled and unlabeled data causes the former to dominate, which severely degrades pseudo-label quality. To address this challenges, we propose a novel remote sensing (RS) $\text{S}^4$ method via unified flow with feature memory bank (UFFM). Specifically, UFFM comprises two key innovations: unified flow (UF) and feature memory bank (FMB). The UF is a new training flow that generates less biased pseudo-labels by combining an external visual foundation model (VFM) with an RS domain teacher, and jointly optimizes labeled and pseudo-labeled data under a unified training objective. The FMB is a novel memory module for $\text{S}^4$ that dynamically updates class-specific features during training and reduces the feature discrepancy between labeled and unlabeled data through class-feature alignment. To verify the effectiveness of our model, we conduct extensive experiments on RS datasets. The experimental results show the superiority of our method over SOTA $\text{S}^4$ methods. Moreover, the results demonstrate the effectiveness of our contributions in bridging the optimization and feature representation gap between labeled and unlabeled data. Our code is released at \href{https://github.com/wangshanwen001/RS-UFFM}{https://github.com/wangshanwen001/RS-UFFM}.
\end{abstract}


\section{Introduction}
\begin{figure}[!t]
\centering
\includegraphics[width=3.5in]{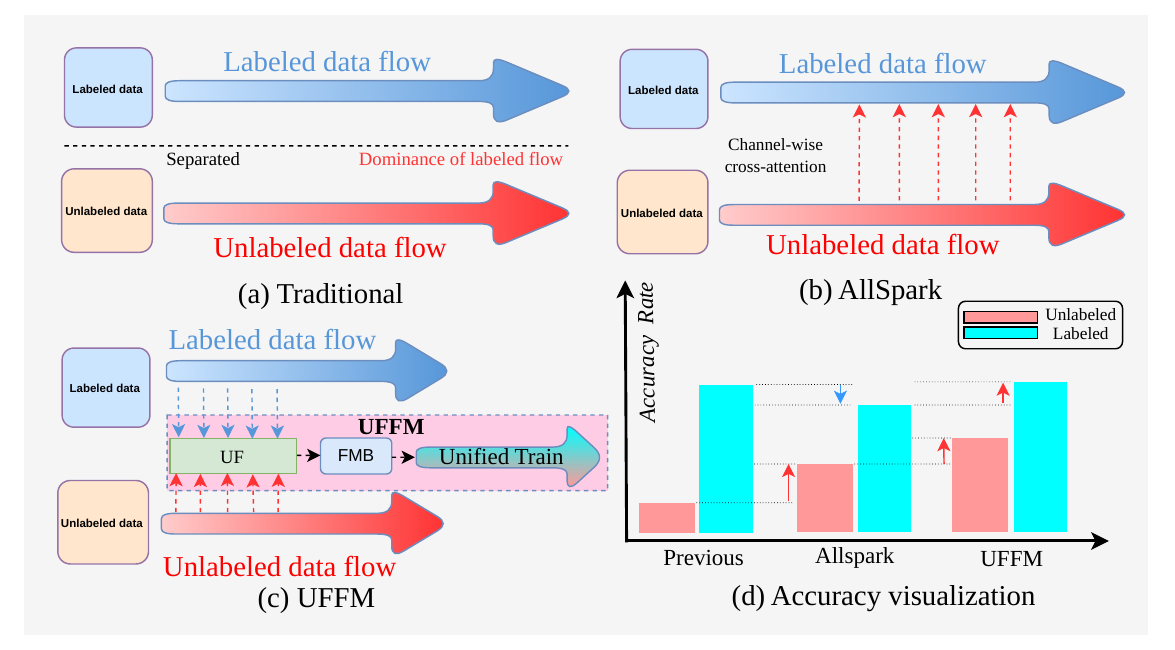}
\vspace{-0.5cm}
\caption{(a), (b) and (c) Comparison between the training data flows of previous methods, AllSpark and ours UFFM. (d) Comparison between the accuracy rate of previous methods, AllSpark and ours UFFM.}
\label{fig: fg1}
\vspace{-0.5cm}
\end{figure}

\IEEEPARstart{S}{emantic} segmentation has shown great potential in applications such as flood monitoring\cite{zhu2026foundations}, precision agriculture\cite{hong2026foundation}, ecosystem assessment\cite{wang2025accurate}, and urban planning\cite{zhou2026transformer}. However, manually labeling extensive RS datasets is highly time-consuming and labor-intensive\cite{xue2025egpo, 11126950,11134807}. Semi-supervised semantic segmentation ($\text{S}^4$) has attracted significant attention in the remote sensing (RS) community by leveraging a small set of labeled samples alongside a wealth of unlabeled data \cite{lv2026s5, sun2025rsprotosemiseg, han2025difference}. Several RS $\text{S}^4$ models address issues like high inter-class similarity, long-tailed distributions, and low-quality pseudo-labels\cite{zhang2025more, huang2024decouple, 11612938}. Nevertheless, training labeled and unlabeled data independently prevents effective feature interaction, causing inconsistent semantic representations. Consequently, the explicitly annotated data dominates optimization, severely compromising pseudo-label quality and aggravating confirmation bias. To address this, AllSpark \cite{wang2024allspark} reconstructs labeled features from unlabeled features using a channel-level cross-attention mechanism. While this approach improves unlabeled data accuracy, it unfortunately degrades labeled data performance, as shown in Fig.~\ref{fig: fg1}(d). This occurs because reconstructing labeled samples from unlabeled features can introduce noise into the discriminative features of labeled data. Specifically, during training, these reconstructed samples align poorly with ground-truth annotations, compromising supervised learning effectiveness. This challenge is even more pronounced in the RS domain by extreme label scarcity.

To address the above challenge, we propose a unified flow with feature memory bank (UFFM). As shown in Fig.~\ref{fig: fg1}, UFFM bridges the gap between labeled and pseudo-labeled data, surpassing traditional $\text{S}^4$ methods. Specifically, UFFM introduces two key innovations: a unified flow (UF) paradigm and a feature memory bank (FMB). Instead of isolating labeled and unlabeled data, UF integrates them through collaborative pseudo-labeling strategy and a data fusion mechanism. It generates less biased pseudo-labels by combining the external knowledge of a vision foundation model (VFM) with the domain-specific expertise of an in-domain teacher. Then, UF mixes these pseudo-labels with labeled data under a unified loss function to establish a cohesive supervision framework. Furthermore, features of the same category across labeled and unlabeled data should exhibit high similarity. To achieve this, we propose a feature memory bank (FMB) to inject long-term category feature memory into $\text{S}^4$ training. The FMB creates and updates a feature prototype for each semantic category and leverages them to evaluate the reliability of pixel-level pseudo-labels. UFFM integrates UF and FMB to mitigate the dominance of labeled data during training. By optimizing labeled and unlabeled samples within a shared feature space, it effectively bridges the gap between labeled and pseudo-labeled data. It yields superior performance across both labeled and unlabeled data compared to existing $\text{S}^4$ baselines (Fig.~\ref{fig: fg1}(d)). In summary, our contributions are as follows:
\begin{enumerate}
\item We propose UFFM, a novel $\text{S}^4$ model for RS, to address the challenge of separated traditional training paradigm, where explicitly annotated data dominates training and ultimately degrades pseudo-label quality.
\item We propose UF that generates less biased pseudo-labels by fusing external VFM knowledge with domain-teacher expertise, and performs a specific  unified training by integrating pseudo-labels with ground-truth labels.
\item We propose the FMB that introduces long-term category feature memory into $\text{S}^4$ training. It optimizes labeled and unlabeled samples within a shared feature space, by updating a feature prototype for each category.
\item Extensive experiments demonstrate the effectiveness of UFFM over state-of-the-art methods. The comprehensive ablation studies highlight our success in bridging the optimization and feature representation gap between labeled and unlabeled data, significantly boosting performance across both data types.
\end{enumerate}

The rest of this article is organized as follows: Section \ref{sec: related work} provides an overview of existing related research. In Section \ref{sec: methods}, we formally propose and analyze our UFFM model. Section \ref{sec: Experiment} presents comprehensive experimental results, including comparative analyses with SOTA methods and ablation studies. Finally, Section \ref{sec: Conclusion} concludes and discusses the article.

\section{related work}
\label{sec: related work}

This section reviews relevant research on $\text{S}^4$ methods, and also summarizes recent progress in $\text{S}^4$ for RS domain. 

\subsection{Semi-supervised Semantic Segmentation}

Semantic segmentation is a foundational image analysis task with widespread application in areas such as land cover classification\cite{11333329}, urban modeling\cite{11373188}, and environmental monitoring\cite{hong2024spectralgpt}. However, conventional supervised methods are constrained by the heavy demand and high cost of pixel-level annotations\cite{11448789, 11417739, fu2025alternating, liu2025confidence}. Consequently, $\text{S}^4$ has attracted growing interest by leveraging large-scale unlabeled data\cite{cheng2025cgmatch, jing2025recursive, zeng2025pick, chen2025conformalsam, yin2026depmatch, yin2025semi}.

In computer vision, classical S4 strategies predominantly rely on consistency regularization, pseudo-labeling, teacher–student frameworks, and adversarial learning\cite{yan2026language, lin2025leo, dang2024progressive}. A representative method is FixMatch\cite{sohn2020fixmatch}, which generates high-confidence pseudo-labels from weakly augmented unlabeled images to supervise strongly augmented views via consistency regularization. UniMatch\cite{yang2023revisiting} extends this paradigm by introducing unified dual-stream perturbations across both image- and feature-level representations, effectively expanding the perturbation space to strengthen weak-to-strong consistency learning. Owing to its superior performance, UniMatch has become a widely adopted baseline. Then, UniMatch\_v2\cite{yang2025unimatch} upgrades the architecture by swapping traditional ResNet backbones for vision foundation models like DINOv2\cite{oquab2023dinov2}. To prevent labeled data from dominating the training process, AllSpark\cite{wang2024allspark} reconstructs labeled features from unlabeled representations using a channel-level cross-attention mechanism.

These methods have achieved remarkable performance in natural-image segmentation and offer valuable design insights for RS $\text{S}^4$ approaches \cite{sun2025beyond, sun2025rsprotosemiseg, xuan2024tsg}. However, RS images present distinct challenges, including substantial scale variations, strong visual similarity across categories, complex backgrounds, and intricate textures \cite{11062866, lu2025uncertainty}. Consequently, directly transferring these general strategies to RS $\text{S}^4$ remains challenging\cite{zhou2025advancing}.

\subsection{Remote Sensing Semi-supervised Semantic Segmentation}
To better accommodate the distinctive characteristics of RS imagery, recent studies have developed $\text{S}^4$ methods specifically tailored to the RS domain\cite{gan2025prior, chen2024category, 10965756, chen2022semi}. For example, Ni et al.\cite{ni2025clr} and Wang et al.\cite{11062866} addressed multi-scale variations through contextual label refinement in the label space and multi-scale uncertainty consistency, respectively. To mitigate the high visual similarity among different classes, Xin et al.\cite{xin2024confidence} combined contrastive learning with a cross-teacher–student attention network. Huang et al.\cite{huang2024decouple} proposed decoupled weighting learning (DWL) to alleviate the adverse effects of inaccurate pseudo-labels and long-tailed class distributions. DWL decouples the predictions for labeled and unlabeled data during training and introduces a rank-based weighting module that adaptively weights pseudo-labels according to their relative confidence within each pseudo-class, thereby improving the reliability of pseudo-label learning. To address insufficient multimodal fusion and limited pixel-level annotations, Li et al.\cite{10965756} proposed Semi-Mamba, which incorporates a semi-supervised Mamba-based cross-modality fusion module to facilitate cross-modal feature interaction in RS imagery. Furthermore, Geng et al.\cite{10679155} proposed KGCSL to address the scarcity of labeled samples in hyperspectral image (HSI) classification. This semi-supervised method integrates multi-scale and multi-directional geometric features extracted using the contourlet transform, enabling accurate HSI classification under limited-label conditions.

With the rapid advancement of large-scale foundation models, vision–language models (VLMs)\cite{yang2025qwen3} and visual foundation models (VFMs)\cite{ravi2025sam} have increasingly been incorporated into RS $\text{S}^4$. For example, SemiEarth\cite{11612938} is the first approach to leverage VLMs to improve pseudo-label quality, thereby enhancing the performance of $\text{S}^4$ on RS imagery. Similarly, Song et al.\cite{11563554} employed multiple VFMs as teachers and introduced a distillation-and-fusion mechanism to guide the training of an RS $\text{S}^4$ framework.

However, existing RS $\text{S}^4$ methods remain largely confined to conventional architectural paradigms. These methods often prioritize increasingly complex framework designs while overlooking the optimization imbalance between labeled and unlabeled data. Consequently, the supervised pathway with labeled samples dominates the training process, resulting in low-quality of the pseudo-labels generated for unlabeled data. To address this limitation, we propose UFFM, which bridges the gap between labeled and pseudo-labeled data.

\section{METHODS}
\label{sec: methods}

This section is organized as follows. Section \ref{sec: main framework} describes the main framework, Section \ref{sec: unified flow} introduces the principles of the UF module, and Section \ref{sec: feature memory bank} presents the basic principles of the FMB module.

\subsection{Main Framework}
\label{sec: main framework}

\begin{figure*}[!t]
\centering
\includegraphics[width=7in]{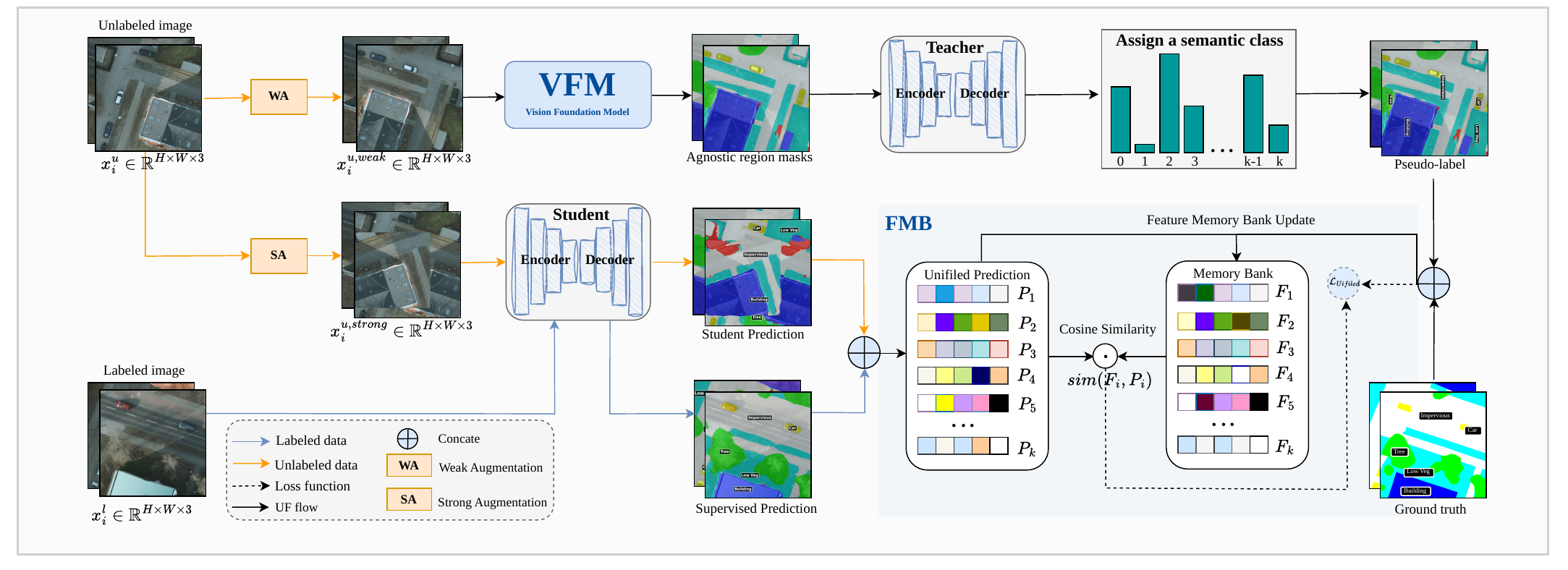}\vspace{-0.3cm}
\caption{Overall architecture of our UFFM model for RS images. Flow paths are color-coded: blue for labeled data, yellow for unlabeled data, black for UF processing, and dashed for loss functions. As shown by the black path in the top row, pseudo-labels in the UFFM model are initially generated by the VFM as class-agnostic masks and subsequently mapped to semantic classes by the teacher. The light blue-gray region highlights the FMB module, which comprises two stages: feature memory bank updating and loss computation, where loss weights are determined via cosine similarity.}
\label{fig: framework}
\vspace{-0.5cm}
\end{figure*}

Semi-supervised learning trains a model using a small set of labeled samples together with a large amount of unlabeled data. As discussed above, conventional strategies that process these two data sources separately fail to fully exploit the potential of unlabeled data. To overcome this limitation, we propose UFFM, a specific $\text{S}^4$ framework illustrated in Fig.~\ref{fig: framework}. Unlike conventional approaches that treat labeled and unlabeled samples as separate training sets, UFFM integrates them through a novel pseudo-labeling strategy and data-fusion mechanism.

We denote the labeled set as $\mathcal{D}^L = \{(x_i^l, y_i^l)\}_{i=1}^{N_L}$ and the unlabeled set as $\mathcal{D}^U = \{x_i^u\}_{i=1}^{N_U}$, where $x_i^l \in \mathbb{R}^{H \times W \times 3}$ and $x_i^u \in \mathbb{R}^{H \times W \times 3}$ denote the labeled and unlabeled images, respectively, with height $H$ and width $W$. $y_i^l \in \{0, 1\}^{H \times W \times K}$ denotes the ground-truth across $K$ classes. $N_L$ and $N_U$ represent the total counts of labeled and unlabeled samples, where typically $N_U \gg N_L$. The standard $\text{S}^4$ loss function is given by:
\begin{equation}
\mathcal L=\frac{1}{N_L}{\sum_{ {i=1}}^{N_L}\mathcal L_{CE}(p_i^l,y^l_i)}+ \frac{1}{N_U}{\sum_{ {i=1}}^{N_U}\mathcal L_{CE}(p_i^{u,s},y_i^{u,t})},
\end{equation}
where the former component denotes the supervised loss on labeled samples and the latter corresponds to the unsupervised loss on unlabeled images. $p_i^l$ is the prediction for labeled input $x_i^l$, and $\mathcal{L}_{\text{CE}}$ represents the cross-entropy loss. While standard $\text{S}^4$ methods process labeled and unlabeled data separately, our proposed UFFM framework trains them jointly through a specially designed weighting strategy. The main loss function is formulated as:


\begin{equation}
\label{eq: loss}
\begin{aligned}
\mathcal L = \frac{1}{N_L+N_U}
\sum_{i=1}^{N_L+N_U}
\mathcal L_{CE}
\Big(
[w_i^l,w_i^u]
([p_i^l,p_i^{u,s}],
[y^l_i,y_i^u])
\Big).
\end{aligned}
\end{equation}
where $w^l_i$ and $w^u_i$ are derived from the FMB module, the details of which are provided in Section~\ref{sec: feature memory bank}. As illustrated in Fig.~\ref{fig: framework}, UFFM unifies supervised and unsupervised learning within a teacher–student framework. For the unsupervised branch, both weak and strong augmentations are applied to unlabeled data to facilitate reliable pseudo-label generation and enhance feature learning robustness. Specifically, weakly augmented samples provide stable representations for the VFM and teacher model to yield high-confidence pseudo-labels, whereas strongly augmented samples expose the student model to broader data variations. The student model is then optimized under the supervision of the pseudo-labels generated by the UF module. The teacher parameters $\theta^t$ are updated via Exponential Moving Average (EMA) from student parameters $\theta^s$, i.e., $\theta_t^t= \alpha\theta_{t-1}^t + (1-\alpha)\theta_{t}^s$ where $\alpha$ denotes the decay rate.  However, conventional methods process labeled and unlabeled data independently, allowing labeled samples to dominate optimization and degrade pseudo-label quality. To resolve this limitation, we introduce the UF training flow and FMB module, detailed in the subsequent sections.

\subsection{Unified Flow}
\label{sec: unified flow}

This section details the principle of the proposed UF module. Unlike traditional paradigms that process labeled and unlabeled data independently, UF introduces a joint pseudo-labeling and data-fusion strategy. To generate less biased pseudo-labels, UF leverages cross-domain collaboration between a VFM and an in-domain teacher network. This strategy leverages the external knowledge encoded in the VFM to mitigate biases inherited from the labeled data while effectively exploiting the teacher model’s domain-specific knowledge. Specifically, for each unlabeled image, the VFM first generates a set of class-agnostic regions. Each region is then assigned the dominant class predicted by the teacher model, and its reliability is assessed based on class consistency and the teacher’s mean confidence. Only regions that satisfy both reliability thresholds are retained as VFM-assisted pseudo-labels. For pixels not covered by reliable VFM regions, we use the teacher predictions that satisfy the conventional confidence threshold. Combining these two sources yields the final pseudo-label map for each unlabeled training sample, thereby reducing its dependence on biases inherited from the labeled data. This process is formally defined as follows. First, the weakly augmented unlabeled image $x^{u,weak}_i$ is fed into the VFM to generate a set of class-agnostic region masks:
\begin{equation}
M_i = \{m^1_i, m^2_i, ... , m^j_i\} = \text{VFM($x^{u,weak}_i$)},  
\end{equation}
where $m^j_i$ denotes the $j^{th}$ class-agnostic mask region generated by the VFM for unlabeld image $i$. The teacher model yields the class probability distribution for each pixel:
\begin{equation}
p^{u,t}_i(x^{u,weak}_i) = Softmax(Teacher(x^{u,weak}_i)).  
\end{equation}
The predicted class $y^{u,t}_i$ and confidence score $c^k_i$ of the teacher for each pixel are given by:
\begin{equation}
y^{u,t}_i = \mathop{\arg\max}_{k \in [1,K]}(p^{u,t}_i(x^{u,weak}_i)),
\end{equation}

\begin{equation}
c^k_i=\mathop{\max}_{k \in [1,K]}(p^{u,t}_i(x^{u,weak}_i)),
\end{equation}
where k indexes the classes. For each region $m$ generated by the VFM, a semantic class is assigned based on the teacher's pixel-level predictions within that region. Specifically, the majority class occupying the largest proportion of the region is determined as:
\begin{equation}
\label{eq: VFM_label}
y^{V}_i(m) = \mathop{\arg\max}_{k \in [1,K]}\sum_{m\in M}{(y^{u,t}_i (m)=k)}.
\end{equation}
We calculate both the class consistency and the average confidence within the target region. This design stems from the rationale that significant discrepancies between the segmentation predictions of the VFM and the domain-specific teacher model signify low inter-model consistency. It indicates that either the VFM produces inaccurate boundaries or the teacher model predicts incorrect semantic categories. Consequently, a region is regarded as a reliable VFM pseudo-label only when both the consistency and confidence thresholds are satisfied.

\begin{equation}
\gamma^m_i = \frac{\sum_{m\in M}{(y^{u,t}_i (m)=y^{V}_i(m))}}{|M|},
\end{equation}
\begin{equation}
c^m_i = \frac{1}{|M|}\sum_{m\in M}c^k_i,
\end{equation}
\begin{equation}
Q^m_i = (1*(\gamma^m_i>=\tau_{\text{consistency}}))*(1*(c^m_i>=\tau_{\text{conf}})),
\end{equation}
where $\tau_{\text{consistency}}$ represents the consistency threshold between the RS teacher and the VFM, and $\tau_{\text{conf}}$ denotes the pseudo-label confidence threshold. Consequently, the pixel-level pseudo-labels are defined as follows.

\begin{equation}
\label{eq: VFM_label2}
y^V_i=
\begin{cases}
y^V_i(m), & m\in M, Q^m_i=1 \\
0, & otherwise
\end{cases},
\end{equation}
where $0$ denotes unreliable pixels. If a VFM region is classified as unreliable, the method reverts to standard teacher pseudo-labels.

\begin{equation}
\label{eq: teacer_label}
y^{u,t}_i = \begin{cases}
y^{u,t}_i (m), &  m\in M, c^k_i>=\tau_{\text{conf}} \\
0, & otherwise
\end{cases}.
\end{equation}
Unreliable pixels from the teacher model are simply discarded during training. By combining Equations (11) and (12), the final UF pseudo-labels $y^u_i$ are unified as follows:
\begin{equation}
\label{eq: uf_label}
y^u_i=
\begin{cases}
y^V_i(m), & m\in M, Q^m_i=1 \\
y^{u,t}_i (m), &  m\in M, c^k_i>=\tau_{\text{conf}} \\
0, & otherwise
\end{cases}.
\end{equation}
We separately weight the generated pseudo-labels $y^u_i$ and the ground truth $p^l_i$ using the FMB module. Finally, the weighted pseudo-labels and ground-truth labels are combined to compute the loss according to Eq. \ref{eq: loss}, completing the UF pipeline. UF combines a VFM with the teacher model to yield less biased pseudo-labels, establishing a unified form of supervision with the ground truth. The next section details the principles and implementation of the FMB module.

\subsection{Feature Memory Bank}
\label{sec: feature memory bank}

\begin{figure}[!t]
\centering
\includegraphics[width=3.65in]{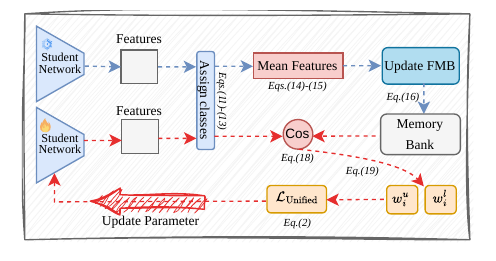}\vspace{-0.3cm}
\caption{The overall workflow of our FMB. It mainly consists of two stages: the memory bank update stage (the red lines) and the loss computation stage (the blue lines).}
\label{fig: FMB}
\vspace{-0.5cm}
\end{figure}

The FMB is a novel memory module for updating class feature prototypes in $S^4$, reducing the feature discrepancy between labeled and unlabeled data through class-feature alignment. The core principle of FMB is that samples from the same category, whether labeled or unlabeled, should exhibit similar feature representations. The overall workflow of FMB is illustrated in Fig. \ref{fig: FMB}. FMB consists of two main stages: memory bank updating (indicated by red lines) and loss computation (indicated by blue lines). The update stage maintains a set of class prototypes within the memory bank to capture and preserve shared characteristics across both labeled and unlabeled data. In the loss computation stage, the model's learned features are compared against these prototypes. By aligning representations with prototypes that encode common characteristics, the proposed approach optimizes labeled and unlabeled samples within a unified feature space, ultimately preventing labeled data from dominating the training process.

During the FMB update stage, which updates the class prototypes, the student encoder is kept frozen to encode image features. FMB applies the same update mechanism to both labeled and unlabeled data. For each class $k \in \{1, \dots, K\}$, we update a class prototype $F_k$. Let $\Omega_k$ denote the set of pixels belonging to class $k$ within a training batch. To determine class membership, unlabeled pixels rely on the pseudo-labels $y^u_i$ provided by UF, whereas labeled pixels use ground-truth annotations. Accordingly, the mean feature vector of class $k$ within a training batch is given by:

\begin{equation}
\label{eq: FMB1}
f_k= \frac{1}{|\Omega_k|} \sum_{i\in \Omega_k}f_i,
\end{equation}
where $f_i$ denotes the feature vector of pixel $i$, while $f_k$ represents the mean feature vector of all pixels belonging to class $k$. Before updating the FMB prototypes, we perform normalization on $f_k$ as follows:
\begin{equation}
\label{eq: FMB2}
f_k= \frac{f_k}{||f_k||_2}.
\end{equation}
The class prototypes in FMB are updated using an EMA mechanism as follows:
\begin{equation}
\label{eq: FMB3}
F_k^t= \alpha F_k^{t-1} + (1-\alpha)f_{k}, 
\end{equation}
where $\alpha$ is the EMA decay and $t$ is training step. Through this mechanism, the class prototypes in FMB are updated to preserve the shared characteristics between labeled and unlabeled data. Next, we describe how FMB derives the loss weights $w^l_i$ and $w^u_i$ in Eq. \ref{eq: loss} for labeled and unlabeled pixels, respectively. 

Having established the updated class prototypes in the FMB module, we next detail how they are utilized during loss computation, as highlighted by the red paths in Fig. \ref{fig: FMB}. Taking unlabeled data as an example, FMB receives both the UF pseudo-labels and the student network's predictions as input. We first use the UF pseudo-label $y^u_i$ to identify the corresponding class prototype. Then, we compute the cosine similarity between the student network's predictions and their corresponding class prototypes as follows:
\begin{equation}
p^{u,s}_i = Student(x^{u,strong}_i),
\end{equation}
\begin{equation}
\label{eq: FMB_compute}
cos^u_i = \frac{p^{u,s}_iF_k}{||p^{u,s}_i||_2 ||F_k||_2}.  
\end{equation}
The resulting cosine similarity values, $cos^u_i$, fall within the interval $[-1, 1]$. To guarantee that final loss function yields non-negative values, $cos^u_i$ is linearly transformed into $w^u_i$ according to:
\begin{equation}
\label{eq: FMB_compute}
w^u_i = \frac{cos^u_i+1}{2}.  
\end{equation}
The value of $w^u_i$ lies in the range $[0,1]$. Similarly, for labeled data, the ground-truth labels $y^l_i$ and the student model predictions are fed to FMB. Following the same procedure as unlabeled data, we compute the corresponding weight $w^l_i$. Finally, these FMB weights are integrated into the UF loss function, as expressed in Eq.~\ref{eq: loss}.

Algorithm~\ref{alg1} summarizes the core training procedure of UFFM to clarify the execution workflow, with the primary logic outlined in lines 7–17. Here, $D^U$ and $D^L$ denote the unlabeled and labeled datasets for a given epoch, containing $N_U$ and $N_L$ images, respectively. $model\_s$ and $model\_t$ represent the student and teacher networks. Furthermore, $p_i^{u,s}$ and $p_i^{l}$ denote the student's output predictions for unlabeled and labeled samples, respectively. Here, $y_i^{u}$ denotes the final UF pseudo-label, $y_i^{l}$ denotes the ground-truth label, and $\mathcal{L}_{CE}$ represents the cross-entropy loss function.

\label{al: UFFM}
\begin{algorithm}[ht]
\caption{Training procedure of UFFM}
\label{alg1}
\begin{algorithmic}
\State 
\State $ \textbf{Input:} $
\State \hspace{0.5cm}$ D^U=\{(x_i^u)\}_{i=1}^{N_U} $, $D^L=\{(x_i^l,y_i)\}_{i=1}^{N_L} $
\State$\textbf{Output}$: 
\State\hspace{0.5cm}$\Theta$: optimal model parameters
\State 1:\textbf{while} until converge:
\State 2:\hspace{0.75cm}\textbf{for} $x_i^l$ , $x_i^u$  in  $ D^L$ , $ D^U$:
\State 3:\hspace{1.25cm}$x^{u,weak}_i=WeakAugment(x_i^u)$
\State 4:\hspace{1.25cm}$x^{u,strong}_i=StrongAugment(x_i^u)$
\State 5:\hspace{1.25cm}$p_i^{l}=model\_s(x^{l}_i)$
\State 6:\hspace{1.25cm}$p_i^{u,s}=model\_s(x^{u,strong}_i)$
\State 7:\hspace{1.25cm}$y^{u,t}_i=model\_t(x^{u,weak}_i)$
\State 8:\hspace{1.25cm}$M_i = \{m^1_i, m^2_i, ... , m^j_i\} = \text{VFM($x^{u,weak}_i$)}$
\State 9:\hspace{1.18cm}\textbf{for} pixels in $m^j_i$:
\State 10:\hspace{1.45cm} \textbf{if} $Q^m_i=1$: 
\State 11:\hspace{1.8cm} $y_i^u=y_i^V(m)$
\State 12:\hspace{1.45cm} \textbf{else if} $Q^m_i!=1$ and $c^k_i>=\tau_{\text{conf}}$: 
\State 13:\hspace{1.8cm} $y_i^u=y_i^{u,t}(m)$
\State 14:\hspace{1.45cm} \textbf{else}:
\State 15:\hspace{1.8cm} $y_i^u=0$
\State 16:\hspace{1.15cm}\textbf{end for} 
\State 17:\hspace{1.05cm} Update FMB by Eq. \ref{eq: FMB1}-\ref{eq: FMB3}.
\State 18:\hspace{1.05cm} Use FMB to compute $w^l_i$ and $w^u_i$ via Eq. \ref{eq: FMB_compute}.
\State 19: \hspace{1.05cm} $\mathcal L = \frac{1}{N_L+N_U}
\sum_{i=1}^{N_L+N_U}$
\State \hspace{2.35cm} $\mathcal L_{CE} \Big( [w_i^l,w_i^u]
([p_i^l,p_i^{u,s}], [y^l_i,y_i^u]) \Big)$
\State 20:\hspace{1.17cm}Update $\Theta$ via gradient descent on $\mathcal L$
\State 21:\hspace{1.17cm}Save the best checkpoint $\Theta_{best}$
\State 22:\hspace{0.75cm}\textbf{end for}
\State 23:\textbf{return} $\Theta$
\State 24:\textbf{end}
\end{algorithmic}
\end{algorithm}


In this section, we presented the core principles of UF and FMB, providing a thorough analysis of how they bridge the optimization and feature representation gap between labeled and unlabeled data. In the following section, we conduct extensive experiments to demonstrate the effectiveness of our proposed methods.

\section{Experiments}
\label{sec: Experiment}
In this section, we evaluate the proposed UFFM model on benchmark remote sensing datasets. We first conduct comparative experiments against several state-of-the-art methods. Subsequently, we perform comprehensive ablation studies to validate the effectiveness of each proposed module. Our code is released at
\href{https://github.com/wangshanwen001/RS-UFFM}{https://github.com/wangshanwen001/RS-UFFM}.

\subsection{Datasets} 
\subsubsection*{\bf DeepGlobe}
The DeepGlobe land cover dataset \cite{demir2018deepglobe} is widely used for remote sensing image semantic segmentation and land cover mapping. It features a spatial resolution of 0.5 meters and comprises 803 high-resolution satellite images of $2448 \times 2448$ pixels across seven categories, including Urban land, Agriculture land, Rangeland, Forest land, Water, Barren land, and Unknown. To facilitate model training, we cropped the original images into $512 \times 512$ patches, yielding 20,075 images in total. The cropped dataset is split into training, validation, and test sets containing 12,045, 4,015, and 4,015 patches, respectively.

\subsubsection*{\bf ISPRS-Potsdam}
The ISPRS Potsdam benchmark dataset is widely used for semantic segmentation on high-resolution remote sensing images \cite{ISPRS_Potsdam}. It features a spatial resolution of 0.05 meters and consists of 38 large-scale satellite images, each with a size of $6000 \times 6000$ pixels. The dataset covers six land cover categories including impervious surfaces, buildings, low vegetation, trees, cars, and background. For efficient training, the original images are cropped into patches of $512 \times 512$ pixels, yielding 5,472 image patches. These patches are partitioned into training, validation, and test sets using a 6:2:2 ratio, containing 3,283, 1,094, and 1,095 images, respectively.

\subsection{Data Augmentation and Experiment Settings}
We first apply data augmentation to the labeled images using geometric transformations (including scaling, horizontal flipping, vertical flipping, and aspect ratio warping) along with Gaussian blurring. For unlabeled images, weak augmentation consists exclusively of geometric transformations, whereas strong augmentation incorporates photometric transformations, Gaussian blur, and CutMix \cite{yun2019cutmix}.

All experiments are conducted on a single NVIDIA RTX H100 GPU using CUDA v11.7. Within our UFFM framework, SAM 3 \cite{carion2025sam} serves as the external-knowledge VFM model, while both the student and teacher adopt a DINOv2-small \cite{oquab2023dinov2} backbone. To evaluate performance across varying supervision regimes, models are trained for 50 epochs on the ISPRS Potsdam and DeepGlobe datasets using $1\%$, $5\%$, and $10\%$ labeled data partitions alongside the remaining unlabeled samples.
\subsection{Evaluation metrics}
Following the standard evaluation protocol of previous RS $\text{S}^4$ methods\cite{huang2024decouple}, we use mean Intersection-over-Union (mIoU) as the primary metric to assess model performance. For DeepGlobe and ISPRS-Potsdam, unknown and background pixels are excluded from evaluation, and mIoU is averaged over the six and five foreground classes, respectively. The mIoU is calculated as follows.
\begin{equation}
IoU_k = \frac{TP_k}{TP_k+FP_k+FN_k},
\end{equation}
\begin{equation}
mIoU=\frac{1}{K}\sum_{k=1}^{K} IoU_k,
\end{equation}
where $TP_k$, $FP_k$, and $FN_k$ represent true positives, false positives, and false negatives for class $k$, and $K$ is the total number of classes.

\subsection{Quantitative Results compared to SOTA}
\begin{table}[ht]
\caption{Comparison results with SOTA methods on DeepGlobe dataset.\ The best results are highlighted in bold. IoU and mIoU are represented as percentages.\label{tab:DeepGlobe}}
\centering
\scriptsize    
\setlength{\tabcolsep}{2pt} 
\begin{tabular*}{\linewidth}{@{\extracolsep{\fill}} c c c c c c c c c}
		\toprule
		\textbf{Ratio} & \textbf{Model} &\multicolumn{6}{c}{\textbf{IoU}}  & \textbf{mIoU} \\
		\cline{3-8}
		~ & ~ & Urban &  Agr. & Range. & Forest & Water & Barren &  ~\\
		\midrule
		\multirow{10}{*}{\textbf{1\%}}& CCT\cite{ouali2020semi} & 70.86 &  70.64 
         & 11.03 & 62.44 & 28.76 & 27.66 & 45.23 \\
        ~& CPS\cite{chen2021semi} & 80.94 &  70.66 
         & 1.16 & 63.91 & 27.45 & 0.79 & 40.82 \\
		~& LSST\cite{lu2022simple} & 79.35 &  73.41 
         & 21.60 & 60.76 & 30.40 & 25.64 & 48.53 \\
		~& FixMatch\cite{sohn2020fixmatch} & 82.51 &  \textbf{74.10} 
         & 18.79 & 67.65 & 44.72 & 32.74 & 53.42 \\
		~& UniMatch\cite{yang2023revisiting} & 80.54 &  70.72 
         & 20.71 & 65.48 & 34.09 & 9.24 & 46.80 \\
        ~& DWL\cite{huang2024decouple} & 81.66 &  75.40 
         & 21.82 & 67.10 & 63.04 & 35.27 & 57.38 \\
		~& AllSpark\cite{wang2024allspark} & 80.69 &  71.15
         & 19.91 & 65.60 & 62.47 & 31.10 & 55.15 \\
        ~ & SemiVL\cite{hoyer2024semivl} & 81.45 &  72.72 
         & 22.44 & 68.24 & 64.40 & 35.76 & 57.50 \\
         ~ & UniMatch\_v2\cite{yang2025unimatch} & 81.34 &  72.35 
         & 22.63 & 68.14 & 65.16 & 36.34 & 57.66 \\
		~& {SemiEarth\cite{11612938}} & 81.63 &  71.43 
         & \textbf{23.43} & 68.74 & 66.99 & 43.16 & 59.23 \\
        ~&\textbf{Our (UFFM)} & \textbf{82.26} &  73.41 
         & 20.62 & \textbf{70.97} & \textbf{71.52} & \textbf{50.49} & \textbf{61.55} \\
		\midrule
		\multirow{10}{*}{\textbf{5\%}}& CCT\cite{ouali2020semi} & 81.20 &  76.14 
         & 12.38 & 64.05 & 49.88 & 42.97 & 54.44 \\
        ~& CPS\cite{chen2021semi} & 84.15 &  78.67 
         & 11.31 & 71.15 & 57.49 & 43.23 & 57.67 \\
		~& LSST\cite{lu2022simple} & 84.26 &  81.67 
         & 30.71 & 68.25 & 65.62 & 55.16 & 64.28 \\
		~& FixMatch\cite{sohn2020fixmatch} & 85.31 &  82.96 
         & 32.22 & 67.47 & 69.76 & 59.09 & 66.13 \\
		~& UniMatch\cite{yang2023revisiting} & 84.13 &  81.36 
         & 30.69 & 69.83 & 65.84 & 54.38 & 64.37 \\
        ~& DWL\cite{huang2024decouple} & 86.08 &  \textbf{83.43} 
         & \textbf{36.62} & 70.22 & 70.77 & 59.86 & 67.83 \\
		~& AllSpark\cite{wang2024allspark} & 80.88 &  81.23 
         & 31.28 & 66.65 & 67.46 & 57.14 & 64.11 \\
        ~ & SemiVL\cite{hoyer2024semivl} & 85.96 &  73.94 
         & 28.48 & 76.39 & 78.58 & 64.16 & 67.92\\
        ~ & UniMatch\_v2\cite{yang2025unimatch} & 84.24 &  73.12 
         & 29.13 & 76.36 & 78.44 & 62.28 & 67.26 \\
		~& {SemiEarth\cite{11612938}} & 86.20 &  70.60 
         & 30.44 & \textbf{77.40} & 80.61 & 64.33 & 68.26 \\
        ~&\textbf{Our (UFFM)} & \textbf{86.54} &  72.69 
         & 31.92 & 74.98 & \textbf{81.45} & \textbf{64.54} & \textbf{68.69} \\
		\midrule
		\multirow{10}{*}{\textbf{10\%}}&  CCT\cite{ouali2020semi} & 83.22 &  80.80 
         & 29.47 & 70.37 & 63.16 & 49.08 & 62.68 \\
        ~& CPS\cite{chen2021semi} & 85.97 &  82.82 
         & 28.20 & 72.03 & 66.97 & 53.76 & 64.96 \\
		~& LSST\cite{lu2022simple} & 85.53 &  83.14 
         & 36.67 & 71.34 & 70.78 & 57.99 & 67.58 \\
		~& FixMatch\cite{sohn2020fixmatch} & 86.53 &  84.01 
         & 36.57 & 71.26 & 69.88 & 57.38 & 67.60 \\
		~& UniMatch\cite{yang2023revisiting} & 84.88 &  82.75 
         & 34.36 & 69.87 & 66.61 & 53.03 & 65.25 \\
        ~& DWL\cite{huang2024decouple} & 85.46 &  \textbf{83.63} 
         & 38.95 & 72.40 & 70.76 & 60.33 & 68.59 \\
		~& AllSpark\cite{wang2024allspark} & 83.66 &  82.21 
         & ~33.41 & 67.80 & 68.24 & 57.22 & 65.42 \\
        ~ & SemiVL\cite{hoyer2024semivl}& 85.25 &  80.73 
         & 38.70 & 76.14 & 73.64 & 63.33 & 69.63 \\
        ~ & UniMatch\_v2\cite{yang2025unimatch} & 83.16 &  81.53 
         & 37.11 & 75.94 & 75.39 & 66.57 & 69.95 \\
		~& {SemiEarth\cite{11612938}} & 86.70 &  73.26 
         & 40.20 & 78.92 & 76.58 & 67.24 & 70.48 \\
        ~&\textbf{Our (UFFM)} & \textbf{87.27} &  76.65 
         & \textbf{41.15} & \textbf{79.32} & \textbf{77.60} & \textbf{67.25} & \textbf{71.54} \\
		\bottomrule
	\end{tabular*}
     \label{tab_Potsdam}
\end{table}

This section conducts experiments on DeepGlobe and ISPRS-Potsdam datasets, compared to the SOTA methods, including CCT \cite{ouali2020semi}, CPS \cite{chen2021semi}, LSST \cite{lu2022simple}, FixMatch \cite{sohn2020fixmatch}, UniMatch \cite{yang2023revisiting}, UniMatch\_v2 \cite{yang2025unimatch}, DWL \cite{huang2024decouple}, Allspark \cite{wang2024allspark}, SemiVL\cite{hoyer2024semivl}, and SemiEarth\cite{11612938}. Specifically, we report the results under labeled data ratios of $1\%$, $5\%$, and $10\%$ to comprehensively evaluate the effectiveness of our method. For all comparative methods, network configurations strictly follow the default settings reported in their respective papers and public code repositories. Some of the baseline results are reported directly from their original publications.

Tables~\ref{tab:DeepGlobe} and \ref{tab_Potsdam} present the quantitative results on the DeepGlobe and ISPRS Potsdam datasets, respectively. Here, Agr. and Range. are abbreviations for Agriculture and Rangeland, respectively. We can observe that traditional $\text{S}^4$ methods, such as FixMatch and UniMatch, perform poorly. This is because, although they improve the accuracy of unlabeled data in various ways, they overlook the unique domain-specific challenges inherent to remote sensing imagery. Conversely, dedicated RS $\text{S}^4$ approaches like DWL achieves noticeable improvements by incorporating architectures tailored to handle distinct RS characteristics, such as rich multi-scale features. Although VLM-based frameworks like SemiVL and SemiEarth outperform standard $\text{S}^4$ baselines, they remain inferior to the proposed UFFM model. Furthermore, the results show that, on both RS datasets, our proposed UFFM achieves the highest mean mIoU across almost all categories. This is because, although previous $\text{S}^4$ methods improve pseudo-label quality through various methods or filter out low-confidence pseudo-labels, they overlook the degradation of pseudo-label quality caused by the labeled-data dominated training process. In contrast, UFFM introduces a new strategy that effectively addresses this limitation.


\begin{table}[t]
\caption{Comparison results with SOTA methods on ISPRS-Potsdam dataset.\ The best results are highlighted in bold. IoU and mIoU are represented as percentages.\label{tab:Potsdam}}
\centering
\normalsize
\scriptsize         
\setlength{\tabcolsep}{2pt} 
\begin{tabular*}{\linewidth}{@{\extracolsep{\fill}} c c c c c @{\hspace{1.25em}} c c c }
		\toprule
		\textbf{Ratio} & \textbf{Model} &\multicolumn{5}{c}{\textbf{IoU}}  & \textbf{mIoU} \\
		\cline{3-7}
		~ & ~ & Building & \makecell{Low \\ vegetation}  & Tree & Car & \makecell{Impervious \\ surfaces} &  ~\\
		\midrule
		\multirow{10}{*}{\textbf{1\%}}& CCT\cite{ouali2020semi} & 54.48 & 61.28 &48.56&52.95&60.71&55.59 \\
        ~& CPS\cite{chen2021semi} & 59.35 & 69.16 &62.89&59.88&66.33&63.52 \\
		~& LSST\cite{lu2022simple} & 68.74 & 75.24 &54.74&62.09&68.80&65.92 \\
		~& FixMatch\cite{sohn2020fixmatch} & 76.95 & 71.59 &64.71&65.85&72.81&70.38\\
		~& UniMatch\cite{yang2023revisiting} & 76.52 & 70.99 &65.44&66.62&72.64&70.44\\
        ~& DWL\cite{huang2024decouple} & 72.34 & \textbf{77.08}&62.74&62.57&72.22&69.39 \\
		~& AllSpark\cite{wang2024allspark} & 83.70 & 65.92 &59.64&69.77&75.31&70.87 \\
        ~ & SemiVL\cite{hoyer2024semivl} & 84.73 & 67.28 & {58.87} & {72.95} &{77.16} & {72.20} \\
        ~ & UniMatch\_v2\cite{yang2025unimatch} & 84.75 &  67.83 
         & 65.88 & 75.50 & 77.15 & 74.22  \\
		~& {SemiEarth\cite{11612938}} & {86.80} & {71.22} &  \textbf{71.96} &  \textbf{76.11} & {79.01}  & {77.02}  \\
        ~&\textbf{Our (UFFM)} & \textbf{88.17} & 70.87 & 71.37 & 75.87 & \textbf{79.48} & \textbf{77.15} \\ 
		\midrule
		\multirow{10}{*}{\textbf{5\%}}& CCT\cite{ouali2020semi} & 72.90 & 80.25 &64.23&58.32&74.42&70.02\\
        ~& CPS\cite{chen2021semi} & 76.53 & 84.34 &57.98&69.45&75.39&72.74 \\
		~& LSST\cite{lu2022simple} & 69.26 & 84.55 &67.33&67.49&73.86&72.50 \\
        ~& FixMatch\cite{sohn2020fixmatch} & 78.12 & 74.87 &68.89&66.58&75.30&72.75\\
		~& UniMatch\cite{yang2023revisiting} & 78.24 & 73.59 &67.17&66.64&75.07&72.14 \\
        ~& DWL\cite{huang2024decouple} & 74.81 & \textbf{85.64} &66.38&62.99&75.68&73.10 \\
		~& AllSpark\cite{wang2024allspark} & 85.57 & 67.62 &60.61&73.48&77.15&72.88 \\
        ~&{{SemiVL\cite{hoyer2024semivl}}} & {87.58} & {70.37} & {63.59} & {75.06} &{78.92}&{75.10} \\ 
        ~ & UniMatch\_v2\cite{yang2025unimatch} & 83.75 &  76.47 
         & 66.40 & 75.95 &77.19~ & 75.95  \\
        ~& {SemiEarth\cite{11612938}} & {88.51} & 74.45 &  {74.06} &  {78.14} &  {79.87} & {79.01}  \\
        ~&\textbf{Our (UFFM)} & \textbf{88.68} & 73.48 & \textbf{74.85} & \textbf{78.72} & \textbf{80.44} & \textbf{79.23}\\
		\midrule
		\multirow{10}{*}{\textbf{10\%}}& CCT\cite{ouali2020semi} & 73.09 & 83.94 &61.12&60.45&73.06&70.33 \\
        ~& CPS\cite{chen2021semi} & 77.80 & 87.15 &61.12&68.48&75.89&74.09 \\
		~& LSST\cite{lu2022simple} & 70.92 & 86.06 &68.91&70.22&74.89&74.20 \\
        ~& FixMatch\cite{sohn2020fixmatch} & 77.97 & 76.17 &70.09&70.97&76.14&74.27 \\
		~& UniMatch\cite{yang2023revisiting} & 77.34 & 87.75 &70.79&56.65&76.46&73.80 \\
        ~& DWL\cite{huang2024decouple} & 76.37 & \textbf{88.42} &66.54&64.37&77.14&74.57 \\
		~& AllSpark\cite{wang2024allspark} & 86.29 & 69.83 &64.17&75.23&78.31&74.76 \\
        ~&{{SemiVL\cite{hoyer2024semivl}}} & {88.62} & {72.23} & {65.40} & {76.34} &{80.71}&{76.66} \\  
        ~ & UniMatch\_v2\cite{yang2025unimatch} & 84.93 &  78.59  & 68.97 & 75.54 &79.83 & 77.57  \\
        ~& {SemiEarth\cite{11612938}} & {90.59} & 75.44 & {75.01} & {79.64} & {83.24} & {80.78}\\
        ~&\textbf{Our (UFFM)} & \textbf{92.03} & {76.01} & \textbf{76.18} & \textbf{80.08} & \textbf{84.38} & \textbf{81.73} \\
		\bottomrule
	\end{tabular*}
     \label{tab_Potsdam}
\end{table}

\subsection{Visualization Results compared to SOTA}
\begin{figure*}[!t]
\centering
\includegraphics[width=7in]{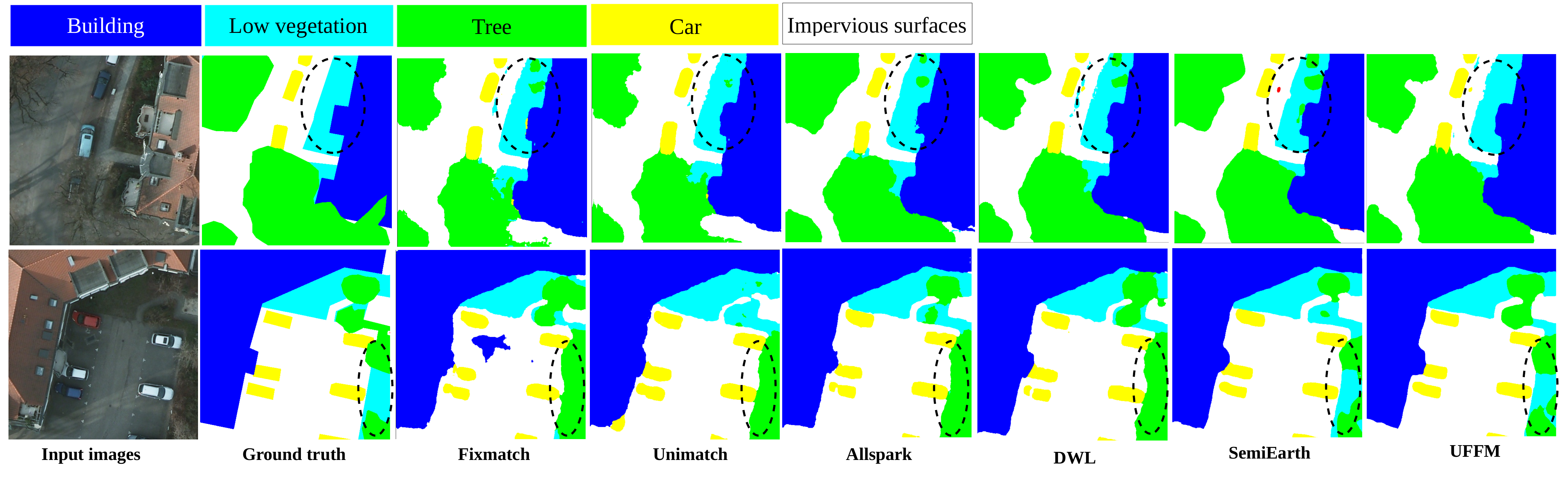}\vspace{-0.2cm}
\caption{Visual comparison of semantic segmentation results with different semisupervised methods on the ISPRS-Potsdam dataset.
\label{fig: Potsdam}\vspace{-0.5cm}}
\end{figure*}

To provide an intuitive comparison, we present visual segmentation results alongside state-of-the-art (SOTA) methods on the RS datasets. Fig.~\ref{fig: Potsdam} illustrates qualitative comparisons on the ISPRS Potsdam dataset. Existing methods exhibit noticeable segmentation errors in complex regions, particularly where class boundaries overlap or intermix. Specifically, as highlighted within the black dashed boxes in Fig.~\ref{fig: Potsdam}, baseline models frequently misclassify the Low vegetation and Tree categories. In contrast, our proposed UFFM produces significantly cleaner boundaries and more accurate predictions.

\begin{figure*}[!t]
\centering
\includegraphics[width=7in]{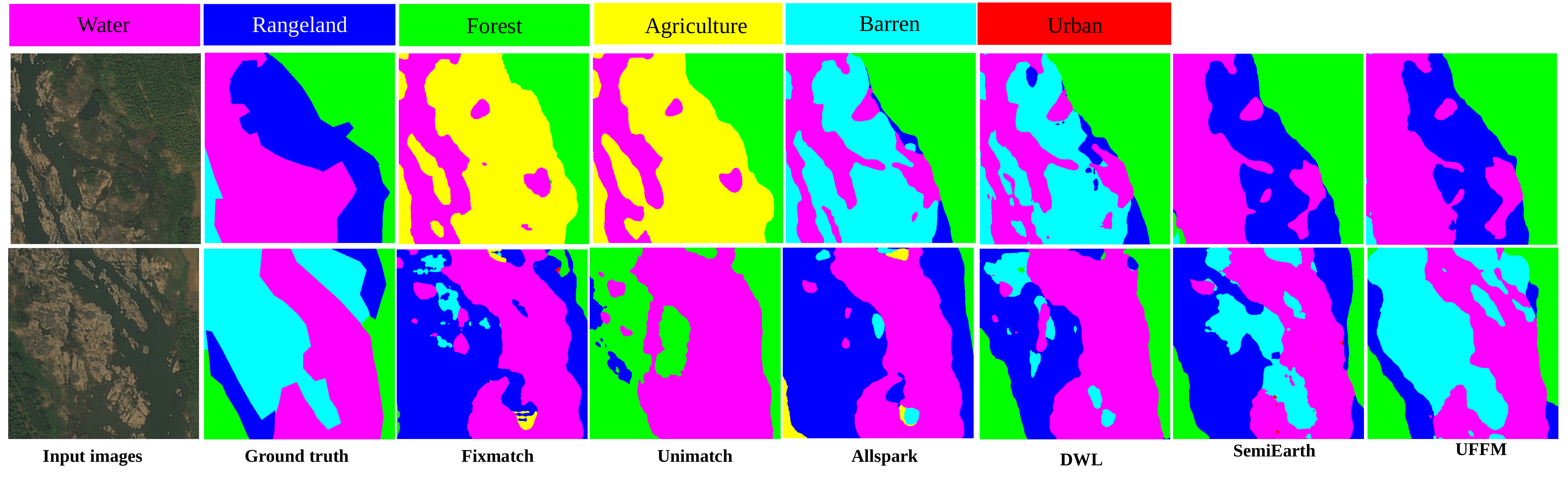}\vspace{-0.2cm}
\caption{Visual comparison of semantic segmentation results with different semisupervised methods on the DeepGlobe dataset.
\label{fig: Goble}\vspace{-0.5cm}}
\end{figure*}

The visualization results  on the DeepGlobe are illustrated in Fig. \ref{fig: Goble}. It can be observed that most models suffer from significant segmentation errors across various scenes. Specifically, in the first row, FixMatch and UniMatch misclassify water and rangeland areas as Agriculture, whereas AllSpark and DWL falsely predict non-existent Barren land. In the second row, FixMatch, AllSpark, and DWL misidentify large regions of Barren land as Water, while UniMatch misclassifies large areas  as Forest. Although SemiEarth yields relatively better predictions, it still suffers from large-scale regional errors. In contrast, UFFM achieves the best segmentation performance, producing precise predictions in fine-grained regions while better preserving structural boundary details. Overall, our proposed model achieves the most robust and accurate performance among all evaluated methods.

\subsection{Ablation Study}

In this subsection, we conduct detailed ablation experiments on UFFM to validate the rationality of our model. We also provide an in-depth analysis and explanation of how our method bridges the optimization and feature representation gap between labeled and unlabeled data.

\subsubsection{Ablation of Components}
To evaluate the individual contribution of each component, we conduct ablation experiments under a 1\% labeled data setting using DINOv2-small as the backbone for both teacher and student networks and SAM 3 as the VFM. We systematically analyze our framework across three configurations: Baseline, UF, and UF + FMB. Specifically, the Baseline model operates without any of our proposed components.

The ablation results are presented in Table~\ref{tab: Ablation of Components}. It can be seen that without the proposed UF and FMB, the performance is poor. Introducing the UF module improves mIoU over the baseline by 3.59\% on the DeepGlobe dataset and 4.23\% on the Potsdam dataset. Incorporating the FMB module provides additional gains, further boosting mIoU by 0.61\% and 1.41\% on DeepGlobe and Potsdam, respectively. These results confirm that UF and FMB significantly enhance $\text{S}^4$ in the RS domain. Furthermore, evaluating performance across labeled and unlabeled subsets reveals that adding UF improves mIoU for both data types, with the joint application of UF and FMB achieving the best overall performance. This validates the necessity and effectiveness of both proposed modules.

\begin{table}[!t]
\caption{Ablation of the Components.}
\label{tab: Ablation of Components}
\centering
\footnotesize

\resizebox{\columnwidth}{!}{%
\begin{tabular}{c|c|c|c|c}
\toprule
\textbf{Dataset} & \textbf{Network} & \textbf{Labeled mIoU} & \textbf{Unlabeled mIoU} & \textbf{mIoU} \\
\midrule

\multirow{4}{*}{DeepGlobe} & Baseline & 91.32 & 56.89 & 57.35 \\
\cmidrule{2-5}
 & UF & 97.83 & 59.74 & 60.94 \\
\cmidrule{2-5}
 & UF+FMB & 98.49 & 60.97 & 61.55 \\
\midrule

\multirow{4}{*}{ISPRS-Potsdam} & Baseline & 88.09 & 70.82 & 71.51 \\
\cmidrule{2-5}
 & UF & 94.50 & 74.48 & 75.74 \\
\cmidrule{2-5}
 & UF+FMB & 96.92 & 77.01 & 77.15 \\
\bottomrule
\end{tabular}%
}
\end{table}

\subsubsection{Hyperparameters Analysis of UFFM}
\begin{figure}[!t]
\centering
\includegraphics[width=3.5in]{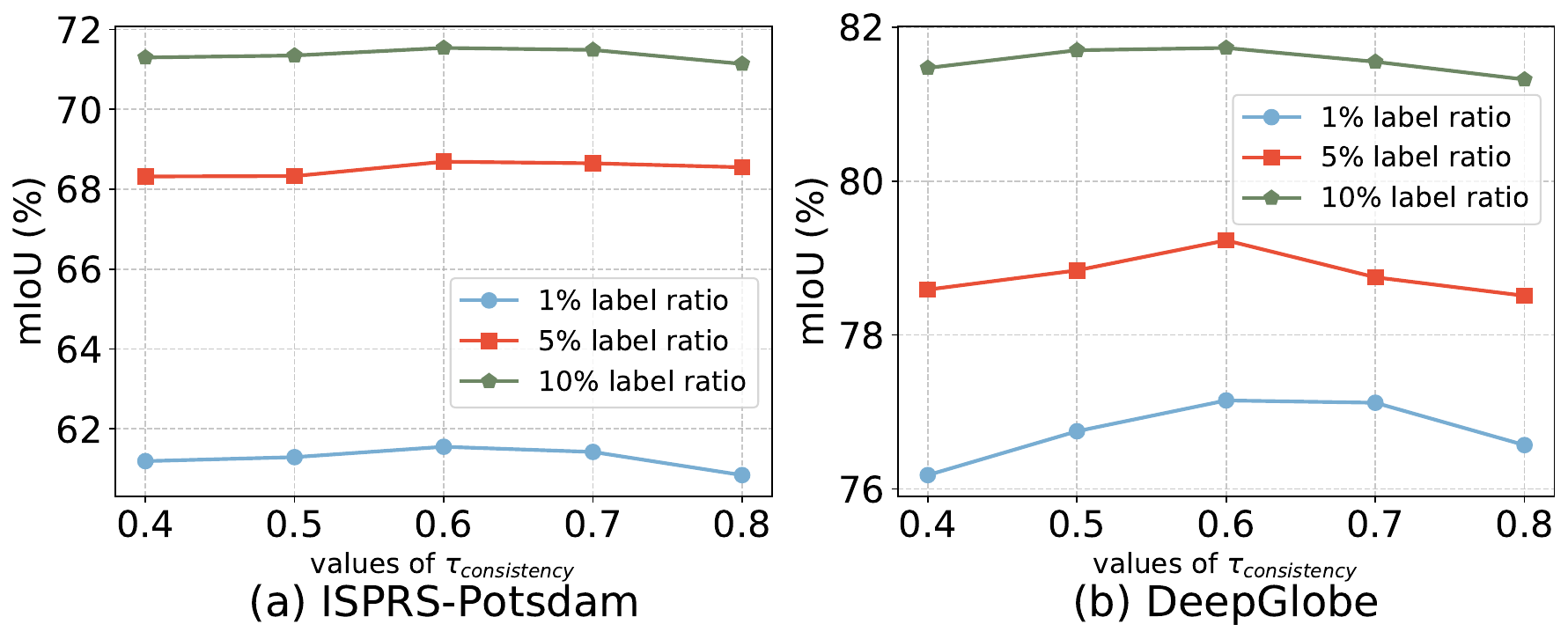}
\vspace{-0.5cm}
\caption{Hyperparameters Analysis $\tau_{\text{consistency}}$ for UF.}
\label{fig: UF1}
\vspace{-0.5cm}
\end{figure}

We perform ablation studies on the key hyperparameters of UFFM, beginning with the consistency threshold, $\tau_{\text{consistency}}$, between the VFM and the domain-specific teacher model. As shown in Fig.~\ref{fig: UF1}, as $\tau_{\text{consistency}}$ increases, the mIoU initially improves before subsequently declining, achieving optimal performance at $\tau_{\text{consistency}} \approx 0.6$. This can be attributed to the fact that, when \(\tau_{\text{consistency}}\) is too low, categories for which the VFM and in-domain teacher model disagree substantially are still accepted. Conversely, when \(\tau_{\text{consistency}}\) is too high, the criterion becomes overly restrictive, causing most pixels to revert to the original teacher model’s predictions.

\begin{figure}[!t]
\centering
\includegraphics[width=3.5in]{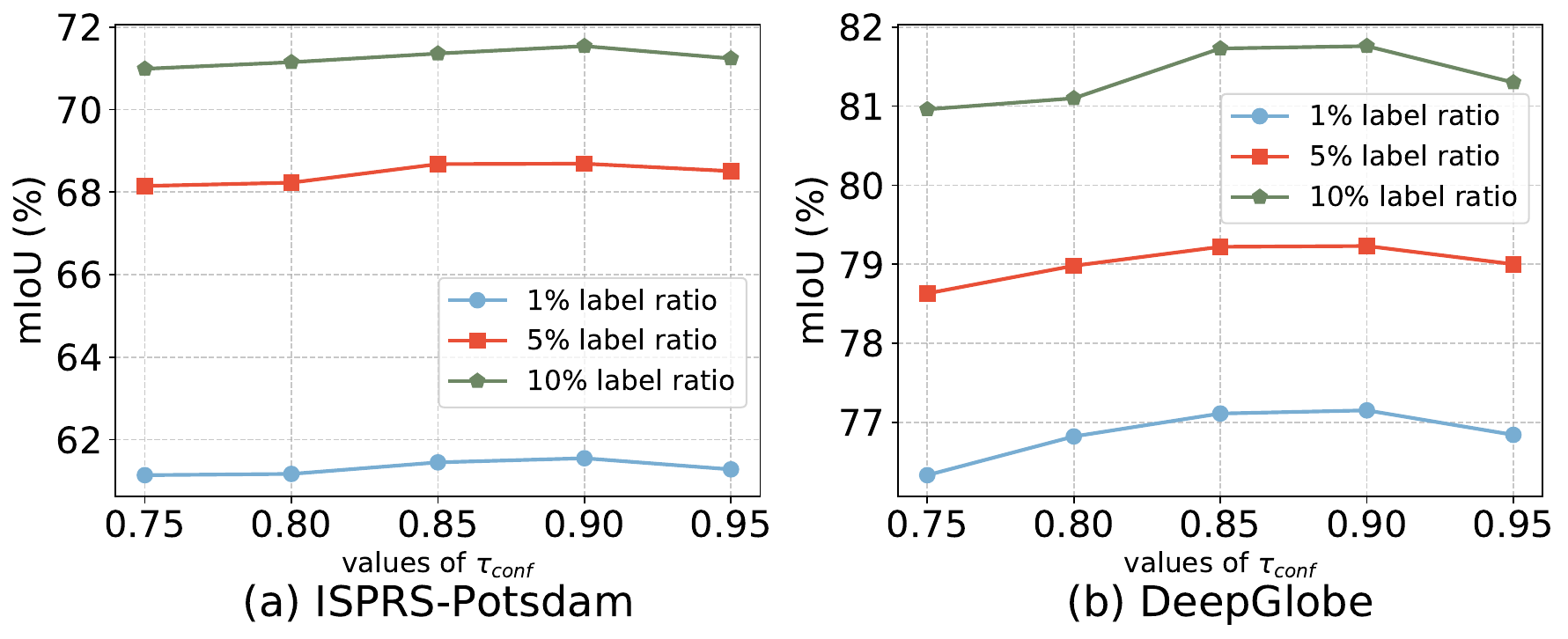}
\vspace{-0.5cm}
\caption{The ablation study of the hyperparameters $\tau_{\text{conf}}$ for UF.}
\label{fig: pseudo_label_threshold}
\vspace{-0.5cm}
\end{figure}

We conducted experiments to evaluate the effect of the pseudo-label confidence threshold $\tau_{\text{conf}}$. As the final safeguard for pseudo-label quality, $\tau_{\text{conf}}$ was set to a relatively high value in our experiments and varied from 0.75 to 0.95. As shown in Fig. \ref{fig: pseudo_label_threshold}, mIoU follows an inverted U-shape trend, i.e., first increasing and then decreasing as $\tau_{\text{conf}}$ rises. This occurs because a lower threshold allows a large number of low-quality pseudo-labels into training, thereby introducing noise and hindering student model optimization. Conversely, a high threshold $\tau_{\text{conf}}$ filters filters out too many pseudo-labels, starving the model of sufficient supervision. Peak mIoU is achieved when $\tau_{\text{conf}}$ is configured between 0.85 and 0.95.

\subsubsection{Backbone Investigation}
We further investigate the impact of backbone architecture by evaluating DINOv2 and DINOv3 backbones ranging from small to large for both the student and teacher models (Table \ref{Ablation of Backbone}). The results show that mIoU on the RS dataset consistently improves as model size increases. These findings confirm that UFFM effectively accommodates backbones of different parameter scales, highlighting its generalizability and versatility of the UFFM network architecture. 

We further observe that DINOv3 consistently outperforms DINOv2 on DeepGlobe, while exhibiting degraded performance on Potsdam. We attribute this discrepancy to the larger patch size adopted by DINOv3. Given the high spatial resolution of Potsdam ($0.05 m$), the larger patch size may discard fine-grained local information, leading to inferior segmentation performance. Considering the trade-off between model size and effectiveness, DINOv2-small proves adequate for most cases with highly stable training. Therefore, it is selected as the default model in this study.

\begin{table}[!t]
\caption{The Backbone Investigation. \label{Ablation of Backbone}}
\centering
\footnotesize
\begin{tabularx}{\columnwidth}{c|C|c|c}
\toprule
		\textbf{Dataset} & \textbf{Backbone} &\textbf{mIoU} & \textbf{Params} \\
		\midrule
		 \multirow{8}{*}{DeepGlobe} &  DINOv2-small &  61.55 & 24.8M \\
        \cmidrule{2-4}  
           ~ & DINOv2-base &  67.38 & 97.5M\\
         \cmidrule{2-4}  
          ~ & DINOv2-large &  68.74  & 335.6M \\
          \cmidrule{2-4}  
          ~ & DINOv3-small &  62.81  &  24.3M \\
           \cmidrule{2-4}  
          ~ & DINOv3-base &  67.92  &  96.6M \\
           \cmidrule{2-4}  
          ~ & DINOv3-large &  69.01  &  334.4M \\
         \midrule
          \multirow{8}{*}{ISPRS-Potsdam} &  DINOv2-small &  77.15 & 24.8M\\
            \cmidrule{2-4}  
          ~ & DINOv2-base  &  80.03 & 97.5M\\
         \cmidrule{2-4}  
          ~ & DINOv2-large  &  80.86 & 335.6M\\
           \cmidrule{2-4}  
          ~ & DINOv3-small &  73.60  &  24.3M \\
           \cmidrule{2-4}  
          ~ & DINOv3-base &  75.67  &  96.6M \\
           \cmidrule{2-4}  
          ~ & DINOv3-large &  78.06  &  334.4M \\
		\bottomrule
\end{tabularx}
\end{table}

\subsubsection{Ablation Study of UF}
We perform an ablation study on the proposed UF to verify its design rationale. Specifically, we decouple the pseudo-label generation process to examine whether first generating anonymous labels with the VFM and then assigning semantic categories using an RS-domain teacher model produces less biased pseudo-labels. We compare three configurations: (1) pseudo-labels generated directly by a conventional RS teacher model (the baseline), (2) pseudo-labels generated directly by the VFM, and (3) pseudo-labels generated using the proposed UF strategy.

The experimental results are presented in Table \ref{tab: Ablation of UF}. The standalone RS teacher model yields the lowest performance. While pseudo-labels generated solely by the VFM marginally outperform those from the RS teacher model, the improvement is slight; despite its large-scale pre-training, the VFM lacks domain-specific RS knowledge. In contrast, our UF framework first leverages the VFM to generate anonymous pseudo-labels and then utilizes the RS teacher model to assign semantic categories, yielding a substantial performance gain. These results confirm that UF successfully capitalizes on the complementary strengths of VFM prior knowledge and RS domain expertise. Furthermore, pseudo-labels generated via UF exhibit reduced bias toward labeled data, ultimately enhancing performance across both labeled and unlabeled sets.

\begin{table}
\caption{Ablation Study of UF.}
\label{tab: Ablation of UF}
\centering
\footnotesize

\resizebox{\columnwidth}{!}{%
\begin{tabular}{c|c|c|c|c}
\toprule
\textbf{Dataset} & \textbf{Network} & \textbf{Labeled mIoU} & \textbf{Unlabeled mIoU} & \textbf{mIoU} \\
\midrule

\multirow{4}{*}{DeepGlobe} & Baseline & 91.32 & 56.89 & 57.35 \\
\cmidrule{2-5}
 & VFM & 92.26 & 57.35 & 58.56 \\
\cmidrule{2-5}
 & Full UF &  97.83 & 59.74 & 60.94 \\
\midrule

\multirow{4}{*}{ISPRS-Potsdam} & Baseline & 88.09 & 70.82 & 71.51 \\
\cmidrule{2-5}
 & VFM & 89.86 & 71.67 & 72.11 \\
\cmidrule{2-5}
 & Full UF & 94.50 & 74.48 & 75.74 \\
\bottomrule
\end{tabular}%
}
\end{table}

\subsubsection{t-SNE Visualizations of Learned Representations}

To evaluate the representations learned by different components of UFFM, we generate t-SNE visualizations across three configurations: the Baseline, Baseline with UF, and the complete model combining both UF and FMB.

 \begin{figure}[]
\centering
\includegraphics[width=3.5in]{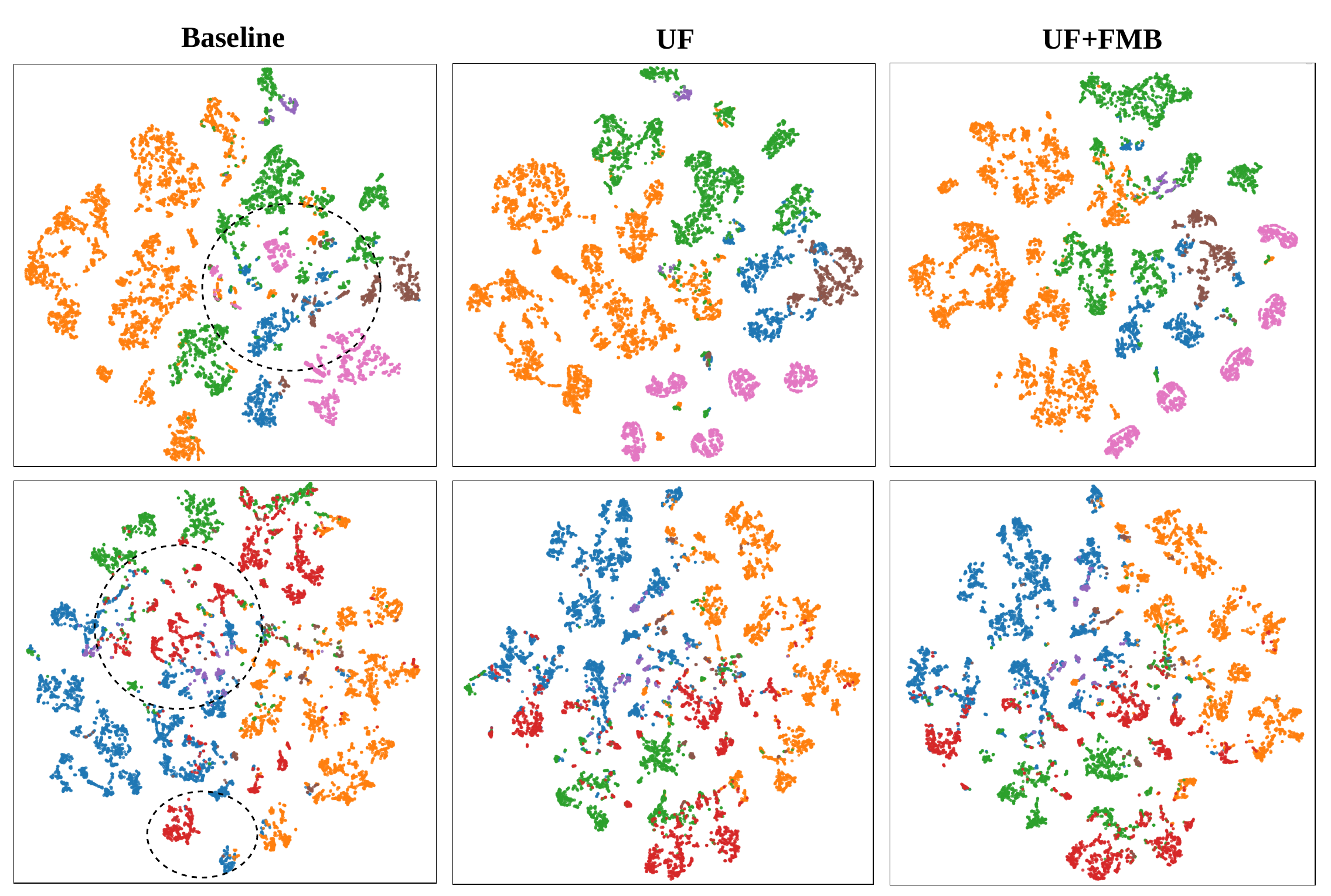}
\vspace{-0.5cm}
\caption{t-SNE Visualizations of Learned Representations.}
\label{fig: tSNE}
\vspace{-0.5cm}
\end{figure}

The experimental results are visualized in Fig. \ref{fig: tSNE}, with different colors denoting features from different classes. The first and second rows show the results on the DeepGlobe and Potsdam datasets, respectively. In the baseline model, features from different classes overlap substantially, while features within the same class are loosely clustered, as highlighted by the black dashed ellipse. These patterns indicate the model’s limited ability to discriminate among classes. After incorporating UF and FMB, the inter-class separation becomes more distinct, and the intra-class feature clusters become more compact, demonstrating substantially improved feature discriminability.

\section{Conclusion}
\label{sec: Conclusion}
In this work, we present UFFM to address the critical issue where independent training on labeled and unlabeled data causes labeled samples to dominate optimization, thereby severely degrading pseudo-label quality. Built upon a unified teacher–student framework, UFFM introduces two novel core innovations: Unified Flow and Feature Memory Bank. UF establishes an innovative training pipeline that synergizes Vision Foundation Models with domain-specific teacher models, successfully mitigating labeled data dominance and improving pseudo-label fidelity. Concurrently, FMB aligns feature representation discrepancies between labeled and unlabeled distributions by memorizing similar features of similar categories. Extensive experiments demonstrate that UFFM achieves state-of-the-art performance in mIoU across remote sensing benchmarks while offering strong interpretability. Comprehensive ablation studies further confirm that UFFM effectively alleviates data bias, enhancing segmentation accuracy simultaneously on both labeled and unlabeled data.

Nevertheless, current semi-supervised RS techniques still exhibit critical bottlenecks. Despite the promising performance achieved by our method in \(\text{S}^4\) domain for RS imagery, its applicability is currently confined to a single downstream task, i.e., semantic segmentation. More broadly, the RS community would benefit from a unified semi-supervised foundation model that can learn effective and transferable representations from limited labeled data while accommodating diverse downstream vision tasks, such as aerial object detection and land-use change detection. Accordingly, future work will investigate the development of a general-purpose semi-supervised foundation model for RS imagery.

Overall, to the best of our knowledge, UFFM is the first framework to address the optimization and feature representation gap between labeled and unlabeled data in the RS $\text{S}^4$ domain. we anticipate that UFFM will establish a foundational benchmark and inspire  future exploration in this promising and emerging direction.

\section*{Acknowledgments}
We would like to express our sincere appreciation to the anonymous reviewers.

\bibliographystyle{IEEEtran}
\bibliography{ref}

\vspace{11pt}
\vfill

\end{document}